# Findings of Factify 2: Multimodal Fake News Detection

S Suryavardan*[1], Shreyash Mishra*[1], Megha Chakraborty[2], Parth Patwa[3], Anku Rani[2], Aman Chadha†[4,5], Aishwarya Reganti[6], Amitava Das[2], Amit Sheth[2], Manoj Chinnakotla[7], Asif Ekbal[8] and Srijan Kumar[9]

[1]*IIIT Sri City, India*
[2]*University of South Carolina, USA*
[3]*University of California Los Angeles, USA*
[4]*Stanford, USA*
[5]*Amazon AI, USA*
[6]*Carnegie Mellon University, USA*
[7]*Microsoft, USA*
[8]*IIT Patna, India*
[9]*Georgia Tech, USA*

#### Abstract
With social media usage growing exponentially in the past few years, fake news has also become extremely prevalent. The detrimental impact of fake news emphasizes the need for research focused on automating the detection of false information and verifying its accuracy. In this work, we present the outcome of the Factify 2 shared task, which provides a multi-modal fact verification and satire news dataset, as part of the DeFactify 2 workshop at AAAI'23. The data calls for a comparison based approach to the task by pairing social media claims with supporting documents, with both text and image, divided into 5 classes based on multi-modal relations. In the second iteration of this task we had over 60 participants and 9 final test-set submissions. The best performances came from the use of DeBERTa for text and Swinv2 and CLIP for image. The highest F1 score averaged for all five classes was 81.82%.

## 1. Introduction

In recent years, automatic fact-checking has become an important problem in the AI community due to the increasing prevalence of fraudulent claims masquerading as declarations of reality. The rapid distribution of news across numerous media sources, particularly on social media

---

platforms, has led to the fast development of erroneous and fake content. The fake news becomes even more harmful during pandemic, elections etc [1, 2, 3, 4]. Uncovering misleading statements before they cause significant harm has become a challenging task. Studies indicate that a large percentage of the population believes that fake news creates uncertainty, while only a fraction feels confident in recognizing bogus news. However, the scarcity of available training data has hindered automated fact-checking efforts.

Significant progress has been made with the release of large datasets like [5, 6, 7], which include extensive claims and contextual metadata. Although these datasets have contributed to research advancements, they were purpose-made and may not capture the patterns present in real-world data effectively. Furthermore, most of the existing datasets focus only on text-based fake news. To address this limitation, Factify 1 [8] introduced a multimodal fact-checking dataset that consists of original samples with no post-processing or manual data creation involved. The dataset includes images, textual claims, and reference textual documents/images, facilitating the exploration of visual cues to enhance fake content detection.

Factify 2 [9], the latest iteration of the Factify dataset, introduced several enhancements and new challenges. Factify 2 expands the dataset with an additional 50,000 instances, encompassing satirical articles that present fake news in a different manner. By incorporating satirical content, it aims to capture the nuanced complexities associated with detecting misinformation in diverse formats. The Factify datasets addresses the limitations of previous unimodal fact-checking research efforts by providing a benchmark for researchers to build and evaluate multimodal fact verification systems. Each data sample in the dataset is labeled with one of five choices: support, no-evidence, refute (both in text and multimodal components), and satirical. The data is collected from popular news channels' Twitter handles in the United States and India, ensuring its relevance to real-world scenarios. Factify 2 builds upon the success of its predecessor by expanding the dataset, introducing satirical articles, and creating new challenges for multimodal fact verification.

This paper presents the findings of the shared task Factify 2, which was organized as part of the workshop at AAAI 2023. The shared task brought together researchers and practitioners to evaluate and compare their approaches in multimodal fact verification using the Factify 2 dataset. Section 2 describes the related work in this domain. We detail the task and the requirements in 3, followed by the participating teams in section 4. The results of all participating teams' models are presented in Section 5. Finally, we summarize the task, discuss further research opportunities, and provide open-ended pointers in Section 7.

## 2. Related Work

**Multi-modal learning:** Researchers have looked into a number of models that incorporate both textual and visual information in the field of multi-modal learning. This includes approaches that range from concatenation of individual embeddings to attention fusion layers. ViLBERT [10] is one such approach that processes both visual and textual inputs separately. They interact through co-attentional transformer layers while extending the BERT architecture. LXMERT [11] builds a large-scale Transformer model with an object relationship encoder, a language encoder, and a cross-modality encoder, for which the model is trained via pre-training tasks like masked language modeling, masked object prediction, etc. to learn intra and cross modality relationships. Learning Joint embedddings using object-word pairs is proposed by [12, 13]. Other notable work includes OFA [14], BLIP [15], ALIGN [16], CLIP[17], etc.

**Unimodal fake news detection:** Many workshops and shared tasks have been conducted on unimodal/text based fake news detection and fact verification [18, 19]. Datasets like LIAR [5], Covid-19 fake news [20], CREDBANK [21] use claims from sources such as social media and PolitiFact and are labeled into fake/real or other fine-grained categories. FEVER [18] provides claims with supporting wikipedia articles that are classified into "Supported", "Refuted" or "NotEnoughInfo". Fever and Feverous[22] use artificial claims that were generated or modified for the task, which may not be useful in real world scenarios. Recently, bert-based methods have been very popular in to tackle text based fake news detection [19, 23, 24]. While unimodal fake news detection might be a good starting point, there is information loss by not utilizing multiple modalities, which may be critical in fake news detection.

**Multimodal fake news detection:** Although under-explored, some of the recent works aim for multimodal fake news detection. The Fakeddit dataset [25] has 1 million samples containing text-image pairs classified into fine-grained labels as well as high-level labels. FakeNewsNet [26], collects data from fact checking websites focused on mitigation and spreading for fake news, and provide social context and dynamic information along with the news. Papadopoulou et al. [27] provide a dataset of 380 verified and debunked videos. Methods to tackle this problem have been proposed in [28, 29, 30, 31]. For a detailed survey on multimodal fake news detection, please refer [32].

**Factify 1:** The previous iteration, Factify 1 shared task [33] at AAAI 2022 was conducted on a multi-modal dataset of having 50k instances. Each data point includes image-text pair for a claim and supporting document, and each claim-document pair is classified into Support, insufficient, neutral. Factify 2 releases additional 50k instances which include data points taken from satirical news.

## 3. Task Details

The Factify 2 task is designed similar to the previous iteration of the task [8]. The task aims to detect fake news with an automated model that is able to classify a given claim based on whether the document entails it or refutes it. Thus, the dataset has 50,000 data points such that each sample has a claim-document pair. Here, the claim is defined as a social media post and the document is an article surrounding the claim. Due to the nature of social media content, the dataset encourages the use of textual and visual features by providing a text and an image for every claim and document. The entailment between the four data sources, namely claim image, claim text, document image and document text, are used to define the categories that the data is to be classified into. This is also shown in Figure 2.

The task description and access to the dataset is available in the Factify 2 task page at https://codalab.lisn.upsaclay.fr/competitions/8275.

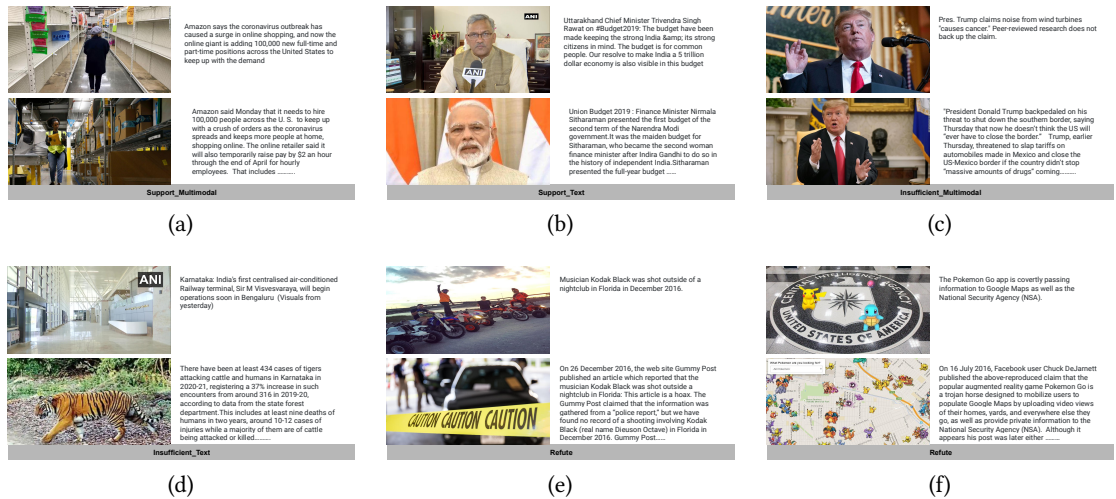

**Figure 1:** Examples for all the 5 categories.

| | | |
|---|---|---|
| Support_Multimodal | Text is supported<br>Similar News | Image is supported |
| Support_Text | Text is supported<br>Similar News | Image is neither<br>supported nor refuted |
| Insufficient_Multimodal | Text is neither supported<br>nor refuted<br>May have common words | Image is supported |
| Insufficient_Text | Text is neither supported<br>nor refuted<br>May have common words | Image is neither<br>supported nor refuted |
| Refute | Fake Claim | Fake Image |

**Figure 2:** This figure shows the five categories the dataset has been divided into along with the relationship between the multi-modal claim and document in each class.

## 3.1. Data

The dataset [9] consists of claim-document pairs and is curated by combining data from Twitter, Fact Checking websites and Satirical news websites 1. The claims were extracted from tweets from Hindustan Times, ANI, ABC and CNN, with their corresponding document text extracted from the news articles linked to the tweets. Based on metrics like textual and image similarity, the collected samples were classified into the Support and Neutral categories. Samples for the refute category were collected from fact checking websites - the fake news was selected as the claim and the article contents were chosen as the corresponding document. Similarly, Satirical websites were scraped for text and image, and as the article supports the satirical i.e. false claim, they were added to the Support category. Satirical headlines were also searched on Google to find news articles that Refuted or were Insufficient for the claim. The train-validation-test split of the 50,000 samples is 70:15:15, with each category having the same number of samples. The train and validation split was provided to the participants and the test set was hidden and used for evaluation. A more in-depth description of the collection process is presented in the data paper for the task [9].

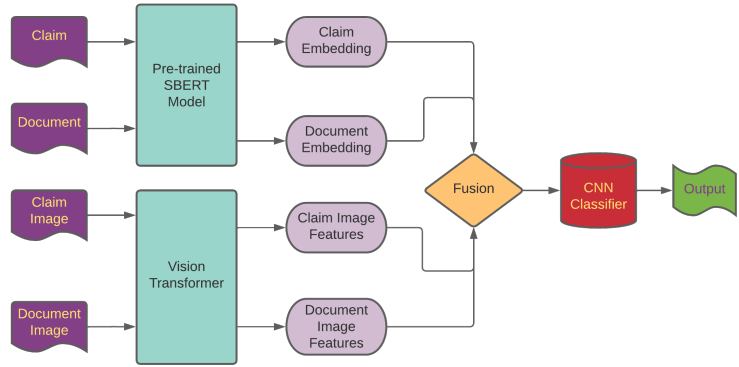

**Figure 3:** This is the baseline model architecture. It extracts multi-modal features using pre-trained models, followed by their fusion and subsequent classification.

### 3.2. Baseline

The baseline for the Factify 2 dataset is presented in [9]. It uses a pre-trained textual and visual feature extraction models before concatenating the features and passing it to a classifier, as shown in Figure 3. The data paper compares few pre-trained models for feature extraction and we finds that the SBERT-mpnet model for text and ViT for image gave the best performance with a F1 score of 64.99%. Please refer to [9] for more details of the baseline.

### 3.3. Evaluation

The Factify 2 dataset has 5 classes, that the data points are to be categorised into: "Support_Text", "Support_Multimodal", "Insufficient_Text", "Insufficient_Multimodal" and "Refute". To compute the performance of the classification models, we use F1 score between the ground truth labels and the predicted labels.

$$F1 = \frac{2 \times Precision \times Recall}{Precision + Recall}$$

$$Precision = \frac{TruePositives}{TruePositives + FalsePositives}, Recall = \frac{TruePositives}{TruePositives + FalseNegatives}$$

We calculate the weighted average F1 score for individual classes and also for all samples together. These scores are computed for all the participating systems and the baseline, as shown in Table 1. The participants were allowed to make 3 submission and the best score out of them was used for the leaderboard. The final F1 score across all classes is used to rank the participants.

## 4. Participating systems

We had over 60 registrations, of which 9 teams provided their final predictions on the test set and 7 teams made paper submissions to the workshop. We provide an overview of the approaches used by the participating teams in this section.

**Triple-Check** [34] propose a model with pre-trained DeBERTa[35] for text and Swinv2[36] for image embeddings, that are combined using a co-attention fusion block. Their novelty is in the use of an adapter to train only a few parameters in their model. Features such as text length, OCR etc. are also used for the final classification.

**INO** [37] use a structure coherence-based approach with components such as textual feature similarity, textual semantic similarity, text length and image similarity. CLIP [17], S-BERT [38] and the ROUGE [39] are used for text features and ResNet for image features. These components are used for the final classification through a Random forest classifier.

**Logically** [40] call their architecture a cross-modal veracity prediction model. They use both uni-modal embeddings from Word2Vec and multi-modal embeddings from CLIP for their classification, after passing them through a multi-head attention layer. The document being fed to this model is from a evidence-retrieval stage, where they rank the paragraphs in document and use the top-K passages.

**Zhang** [41] use uni-modal embeddings from pre-trained DeBERTa and ViT, passed to signed attention layers and a feed-forward network, as well as multi-modal embeddings from CLIP for their classification. This defined as three modules, namely a text-semantic feature module, a image-semantic feature module and a text-image correlation module, together referred to as 'UFCC'.

**gzw** [42] experimented with a inter-modality and intra-modality fusion of textual and visual embeddings using the co-attention mechanism for their classification model. They refer to this architecture as Multimodal Attention and Fusion Network (MAFN). They use different pre-trained models within MAFN and make predictions with an ensemble.

**coco** [43] use an embeddings layer to combine individual embeddings from pre-trained DeBERTa for text and DeiT[44] for images. This embedding is then passed to bi-directional hybrid attention mechanism to fuse the claim and document data. The final classification uses an ensemble of two model pipelines.

**Noir** [45] obtain image features from pre-trained CLIP-ViT and text features from multilingual CLIP-ViT. Claim and document embeddings are concatenated before being fed to a multi-headed attention layer. Then they train a MLP for classification with the output features for image and text concatenated.

## 5. Results

| Rank | Team | Support_Text | Support_Multimodal | Insufficient_Text | Insufficient_Multimodal | Refute | Final |
|---|---|---|---|---|---|---|---|
| 1 | **Triple-Check** [34] | **82.77%** | **91.38%** | 85.19% | **89.22%** | **100.0%** | **81.82%** |
| 2 | INO [37] | 81.24% | 90.03% | **88.81%** | 85.23% | 99.93% | 80.80% |
| 3 | Logically [40] | 80.38% | 90.51% | 84.39% | 85.63% | 98.51% | 78.97% |
| 4 | zhang [41] | 76.64% | 87.85% | 81.61% | 87.93% | 99.93% | 77.42% |
| 5 | gzw [42] | 78.49% | 86.32% | 81.42% | 83.27% | **100.0%** | 76.05% |
| 6 | coco [43] | 77.25% | 86.49% | 81.52% | 83.00% | **100.0%** | 75.70% |
| 7 | Noir [45] | 77.10% | 87.26% | 78.49% | 81.56% | 99.70% | 74.52% |
| 8 | Yet | 70.75% | 82.63% | 78.59% | 71.90% | **100.0%** | 69.09% |
| - | BASELINE | 50.00% | 82.72% | 80.24% | 75.93% | 98.82% | 64.99% |
| 9 | TeamX | 58.22% | 70.91% | 53.66% | 55.56% | 69.79% | 45.62% |

**Table 1**
Leaderboard for the Factify task. The teams are ranked by the overall weighted average F1 score (`Final` column). We also report the category-wise weighted avg F1 score for each team.

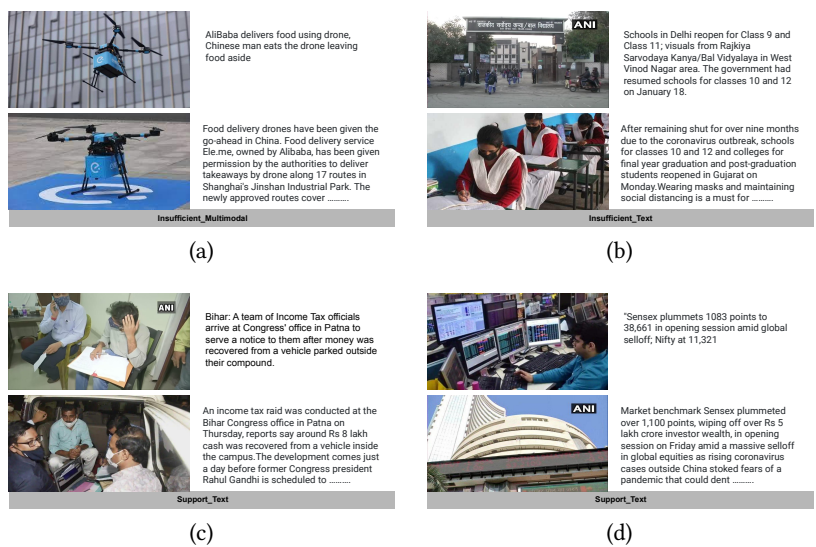

**Figure 4:** Examples from the dataset that were incorrectly predicted by all participants of the task. The Support_Text category has the highest number of such samples, followed by Insufficient_Multimodal, Support_Multimodal, Insufficient_Text and finally Refute in descending order.

For Factify 2, the participants were allowed to make a maximum of three submissions on the test set. The Final F1 score was used to decide which of the three prediction files is the best performing for each participant. Based on that criteria, we present the results of the 9 teams that made test-phase submissions in Table 1. The table shows the weighted average F1 score for the 5 individual classes as well as for the overall test set. All teams except one, improved on the baseline by a minimum of 6.3%. The best performing team is Triple-Check [34] with a Final Score of 81.82%, which is about 26% higher than the baseline. No single team performed better than other team in all categories, which shows that the problem is challenging and needs further research attention.

Despite adding the satire news articles, the results for the Refute category are very high, similar to the Factify 1. This may be due to the fact that the document collected from of fact checking websites are different to the writing in formal news articles. While only having a marginal difference, Support_Text has the lowest scores compared to other categories. This is further emphasised by the observation that, of samples that were predicted incorrectly by all participants, Support_Text has the highest occurrence, while Refute followed by Insufficient_Text occur the least. There are some confusing data points where all the systems failed to predict the correct class. Some such examples are displayed in Figure 4. From the first example, we can see that the document is unrelated to claim, but both of them have a similar drone image, hence leading to confusion. Similarly, in the second example, both the document and claim are about schools re-opening but in different locations, which the models failed to capture.

## 6. Conclusion and Future Work

In this paper, we outline and discuss the Factify 2 shared task on multi-modal fact verification. The participants utilized various techniques for text embeddings, including DeBERTa, CLIP, S-BERT, ROUGE and Word2Vec. The image embeddings were extracted through Swinv2, ResNet, CLIP, ViT and DeiT. Similar to Factify1, ensemble techniques were popular, and some teams opted to use multiple embeddings to capture features. The shared task described in this work seeks to identify fake news, but we are far from our goal. No single system could excel in all categories and there are few examples where all the systems failed, which highlights the challenges of the task. One possible direction for further research includes using synthetic fake news data that matches the general data distribution, thus adding complexity to the refute category.